\documentclass{article}

\usepackage[preprint]{corl_2022} 

\usepackage{graphicx}
\usepackage{amsmath}
\usepackage{listings}
\usepackage{comment}

\begin{document}

\title{Comparison of Model-Free and Model-Based Learning-Informed Planning for PointGoal Navigation}

%

\author{
  Yimeng Li*, Arnab Debnath*, Gregory J. Stein, and Jana Ko{\v{s}}eck{\'a}\\
  Department of Computer Science\\
  George Mason University 
  United States\\
  \texttt{yli44, adebnath, gjstein, kosecka@gmu.edu} \\
}

\maketitle
\def\thefootnote{*}\footnotetext{These authors contributed equally to this work}
\def\thefootnote{\arabic{footnote}}

\begin{abstract}
In recent years several learning approaches to point goal navigation in previously unseen environments have been 
proposed. 
They vary in the representations of the environments, problem decomposition, and experimental evaluation.
In this work, we compare the state-of-the-art Deep Reinforcement Learning based approaches with Partially Observable Markov Decision Process (POMDP) formulation of the point goal navigation problem. 
We adapt the (POMDP) sub-goal framework proposed by~\cite{stein2018learning} and modify the component that estimates frontier properties by using partial semantic maps of indoor scenes built from images' semantic segmentation. 
In addition to the well-known completeness of the model-based approach, we demonstrate that it is robust and efficient in that it leverages informative, learned properties of the frontiers compared to an optimistic frontier-based planner. 
We also demonstrate its data efficiency compared to the end-to-end deep reinforcement learning approaches. 
We compare our results against an optimistic planner, ANS~\cite{chaplot2020learning} and DD-PPO~\cite{wijmans2019dd}
on Matterport3D~\cite{chang2017matterport3d} dataset using the Habitat Simulator~\cite{savva2019habitat}.
We show comparable, though slightly worse performance than the SOTA DD-PPO approach, yet with far fewer data. 

\end{abstract}


\section{Introduction}
In the point goal navigation task,  a robot is deployed in a previously-unseen environment and navigates to a point goal in the egocentric camera coordinate frame. 
This task has a long history in robotics~\cite{thrun1999minerva} and embodied computer vision, and is also an essential component for many downstream tasks including object rearrangement~\cite{gu2022multi}, or disaster response scenarios~\cite{bradley2021learning}.
With the recent emergence of visually rich simulators~\cite{savva2019habitat} and abundance of real-world environments~\cite{chang2017matterport3d, xia2018gibson}, several learning-based approaches~\cite{wijmans2019dd, chaplot2020neural, georgakis2022uncertainty} have been proposed. 
These learning-driven strategies for point-goal navigation~\cite{wijmans2019dd, partsey2022mapping} are usually model-free, in the form of an end-to-end architecture, and trained via deep reinforcement learning.
These policies are learned directly from images or intermediate representations of indoor scenes and are data-intensive to train, requiring millions or billions of frames of training data. 



Instead of learning policies end-to-end directly from observations, another class of RL-based approaches~\cite{chaplot2020learning, ramakrishnan2020occupancy, georgakis2022uncertainty} adopts a hierarchical approach. 
A high-level planner that predicts the waypoints on the map in the unknown area is approximated using a neural network and trained through reinforcement learning.
Local navigation to predicted waypoints is handled by traditional path finding and following algorithms.
Despite some empirical robustness, the waypoints are still predicted via a black-box neural network, and there is no guarantee that they are reachable.

Recently, Stein et al.~\cite{stein2018learning} proposed a learning-augmented model-based approach for point goal navigation. They introduce a high-level abstraction by assigning subgoals to frontiers (boundaries between free and unknown space) and define high-level actions corresponding to traveling to the point goal via a subgoal. 
This abstraction allows for model-based planning over the available frontiers and approximates a complex POMDP as a much simpler Markov Decision Process (MDP).
They use a simple fully-connected neural network to learn the properties of the subgoals from local laser scans.
Conditioned on accurate mapping and reliable low-level navigation, planning via this high-level model-based approach is robust.
However, they only conduct experiments in office-like environments with a robot equipped with a laser sensor, limiting the information the agent can use to make predictions of unseen space.

In this work, we adapt Learning over Subgoals Planner (LSP)~\cite{stein2018learning} and evaluate its performance on a large-scale pseudo-realistic indoor dataset (Matterport3D~\cite{chang2017matterport3d}) with the Habitat simulator~\cite{savva2019habitat}.
Motivated by \cite{ramakrishnan2022poni}, we modify the learning module of LSP (designed initially for a laser range finder) by using visual sensing. 
We use U-Net deep learning architecture to learn the properties of the frontiers from semantic segmentation of images, projected on the semantic map using depth maps. 
We compare our LSP-UNet approach with an optimistic classical planner, a state-of-the-art end-to-end RL navigator (DD-PPO~\cite{wijmans2019dd}), and a modular hierarchical RL-based approach~(ANS \cite{chaplot2020learning}). We evaluate the performance of the LSP-UNet and optimistic planner using both the ground truth semantic map and the semantic map built from sensor observations. 

\section{Related Work}
\noindent \textbf{Embodied Agent Navigation}
Embodied agent navigation~\cite{deitke2022retrospectives} refers to an agent's ability to navigate and interact with its environment in a way similar to how a human would. 
In recent years, there has been a surge of research in embodied navigation thanks to the availability of photo-realistic simulators~\cite{szot2021habitat, habitat19iccv, kolve2017ai2, xia2018gibson} for large-scale indoor environments~\cite{chang2017matterport3d, ramakrishnan2021hm3d, yadav2022habitat}.
The proposed tasks include PointGoal Navigation~\cite{partsey2022mapping}, ObjectGoal Navigation~\cite{georgakis2019simultaneous, georgakis2021learning, pal2021learning, chaplot2020object}, ImageGoal Navigation~\cite{chaplot2020neural, savinov2018semi, hahn2021no, li2020learning} Vision-and-Language Navigation (VLN)~\cite{krantz2021waypoint, krantz2022iterative}, and Object Rearrangement~\cite{batra2020rearrangement, sarch2022tidee, gu2022multi}.
These efforts mainly aimed to develop learning-based strategies that enable the agent to carry out the tasks mentioned above in novel, previously unseen environments. 
The existing approaches typically use end-to-end RL-based models without explicit memory and train a policy using direct raw sensor measurements~\cite{wijmans2019dd}, or intermediate representations of the visual input~\cite{mousavian2019visual, sax2018mid, shen2019situational} (e.g., depth maps, semantic segmentation, local occupancy maps). 
The traditional non-learning-based approaches typically start with exploration~\cite{yamauchi1997frontier} and SLAM~\cite{cadena2016past} to build a map of the environment, followed by path planning and trajectory following.
On the other hand, modular-based approaches exploiting SLAM~\cite{gupta2017cognitive}, path planning, trajectory following~\cite{kumar2018visual}, and exploration~\cite{li2022learning}.
The adopted LSE approach is a modular-based framework with a mapping module, a high-level path planner, and a local navigator.

\noindent \textbf{End-to-End Point Goal Navigation}
Recent deep reinforcement learning methods~\cite{wijmans2019dd, fang2019scene, ramakrishnan2021habitat} 
learn a policy to predict low-level actions directly from raw RGB-D observations or local occupancy maps to improve navigation performance or exploration coverage. 
These end-to-end RL approaches are usually sample inefficient and require a large number of episodes for convergence. 

\noindent\textbf{Intermediate Goals}
Alternative deep reinforcement learning methods~\cite{chaplot2020learning, ramakrishnan2020occupancy, georgakis2022uncertainty, chen2020learning} use hierarchical models that know global policies that predict long-term goals and use local navigation modules to reach the long-term goal.
However, such methods still choose long-term goals greedily, and these long-term goals are usually located in an unknown area that may not be reachable by the agent.
Instead, Stein et al.~\cite{stein2018learning} use a learning-augmented frontier-based abstraction for point goal navigation. 
The robot is guaranteed to reach the goal, and the learning module requires significantly less training data than the model-free RL approaches under their model-based formalism.

\section{Learning-Augmented Model-Based PointGoal Navigation}
\label{sec:approach}

\begin{figure*}[t!]
\centering
\includegraphics[width=1.0\textwidth]{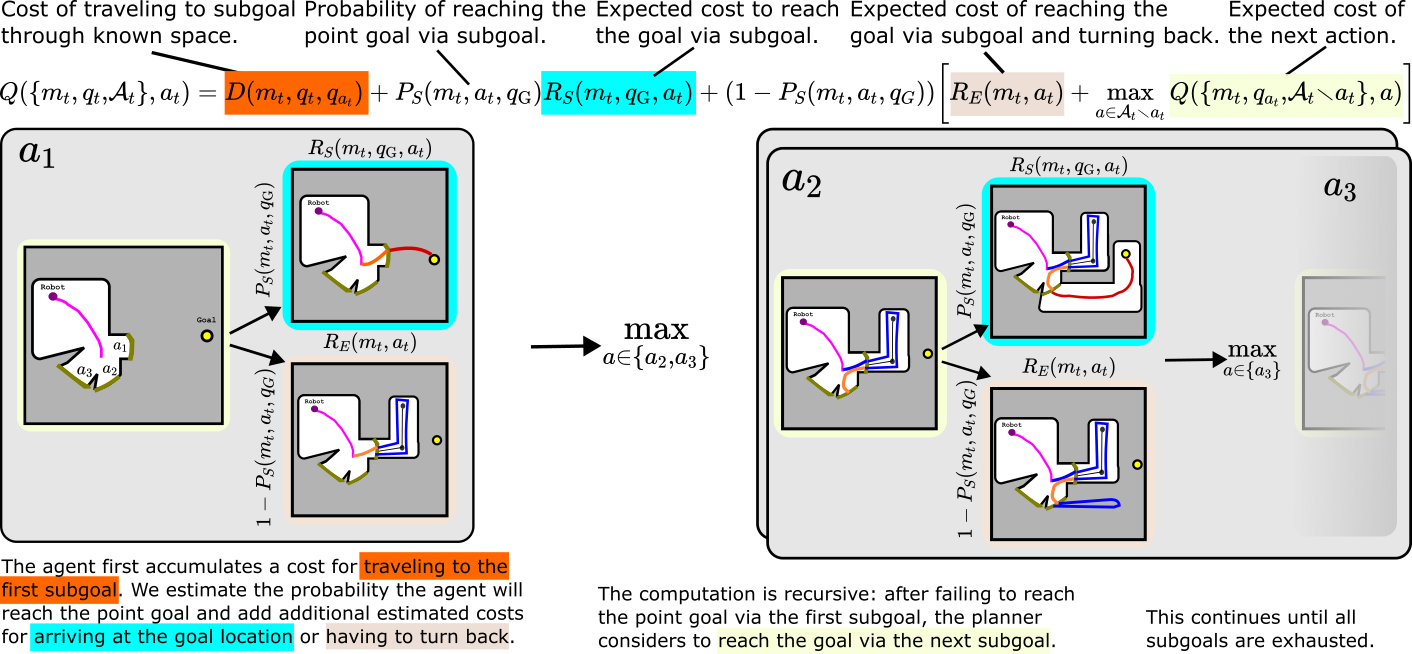}
\caption{
This diagram gives an overview of the LSP-UNet algorithm, which allows us to compute the expected cost of each action in a computationally feasible way for planning through an unknown environment.
Compared to the original LSP~\cite{stein2018learning} approach which estimates the properties of each frontier separately from laser scan, we use a camera sensor and use U-Net to estimate the terms $P_\text{S}$, $R_\text{S}$ and $R_\text{E}$ from the map, thereby introducing prior knowledge about environment regularities into the decision-making procedure.
}
\label{fig:approach}
\end{figure*}

We adapt the Learning over Subgoal Planning (LSP) approach~\cite{stein2018learning} to tackle the task of reaching a point goal in unseen space in a minimum expected distance.
LSP is a high-level planning framework in which navigation in a partially-mapped environment is formulated as a Partially Observable Markov Decision Process (POMDP). LSP introduces an abstraction to simplify the calculation of the expected reward in long-horizon planning.
Under this abstraction, high-level actions correspond to \emph{subgoals}, each associated with travel to and navigation beyond a \emph{frontier}---a boundary between free and unseen space---in an effort to reach the unseen goal.
Under this high-level planning abstraction, a Bellman Equation (Eq.~\ref{eq:dp}) is used to represent the expected cost of executing a high-level action $a_t$ from the belief $b_t$, represented as a three-element tuple consisting of the partially observed map $m_t$, the robot pose $q_t$, and the subgoal set $\mathcal{A}_t$ : $b_t = \{m_t, q_t, \mathcal{A}_t\}$.
The expected cost is defined as:
\begin{equation}
\label{eq:dp}
\begin{split}
Q(\left\{m_t, q_t, \mathcal{A}_t  \right\},a_t) & =  D(m_t, q_t, q_{a_t}) + P_\text{S}(m_t, q_\text{G}, a_t)R_\text{S}(m_t, q_\text{G}, a_t) \\
& + (1-P_\text{S}(m_t, q_\text{G}, a_t))\biggl[R_\text{E}(m_t, a_t)+  \max_{a\in \mathcal{A}_t \setminus a_t} Q(\left\{m_t, q_{a_t},  \mathcal{A}_t \setminus  a_t \right\}, a)\biggr]
\end{split}
\end{equation}

Each action $a_t$, and its corresponding subgoal, is characterized by three properties, each a statistic of the unseen space to which it corresponds: the probability of the subgoal leading to the point goal $P_\text{S}$, expected cost of reaching the point goal via the subgoal $R_\text{S}$, and expected cost of failure (i.e., exploration) in case the subgoal doesn't lead to the point goal $R_\text{E}$.
We train a U-Net~\cite{ronneberger2015u} to estimate the subgoal properties for each subgoal-action, $P_\text{S}$, $R_\text{S}$ and $R_\text{E}$, based on the currently observed map $m_t$ and the relative point goal location $q_\text{G}$.
$D(m_t, q_t, q_{a_t})$ is the cost of traveling from the current location $q_t$ to the centroid of the frontier $q_{a_t}$. Note that this motion occurs entirely in the explored area of the map, so we use Dijkstra's algorithm to calculate this cost.

While LSP is not specific to navigation planning, it relies on frontier abstraction to simplify the decomposition of the problem and calculate the expected reward in long-horizon planning. 
The set of frontiers is usually a small finite set. 
The frontier is a unique abstraction for 2D occupancy maps specific to navigation. 
To extend LSP to other problems, e.g., object rearrangement~\cite{batra2020rearrangement}, suitable simplifying abstractions need to be found. 

\subsection{Hierarchial Navigational Model}
Similar to~\cite{chaplot2020neural,ramakrishnan2020occupancy,georgakis2022uncertainty, garrett2021integrated}, our planner is hierarchical: high-level planning involves selection among frontier subgoals via Eq.~\ref{eq:dp}, while low-level planning involves navigation to the chosen frontier.
Planning with our model to reach a point goal involves two stages:
in the first stage of navigation where the point goal remains in the unknown area of the map, we make a high-level plan to reach the subgoal:
(i) we compute the set of frontiers and the respective action set $\mathcal{A}(b_t)$ given the map $m_t$,
(ii) we run the U-Net to estimate the properties $P_\text{S}$, $R_\text{S}$ and $R_\text{E}$ of each frontier,
(iii) we compute the expected cost for each action according to Eq.~\eqref{eq:dp},
(iv) we choose the action $a_t^*$ that minimizes the expected cost. 
The second stage of navigation involves low-level planning where, 
(v) we compute a motion plan to travel to the centroid of the frontier associated with action $a_t^*$ ($q^*$) and finally
(vi) we select the primitive action that best moves along the computed path and updates the map given the newly observed area.
This process repeats at each timestep until there is a navigable path to the point goal on the explored area of the map $m_t$.


An illustration of our procedure can be found in Fig.~\ref{fig:approach}.

\section{Map and Frontier Computation}
\label{sec:map}
\noindent\textbf{Occupancy and Semantic Mapping}\quad{}
The occupancy map is a metric map of size $m \times m$ where each cell of the map can have three values: 0 (unknown), 1 (occupied), and 2 (free space).
The semantic map is a metric map of size $m \times m$ where each cell's value is in the range $[0, N]$.
Each value from 1 to $N$ corresponds to one of the N object categories, and 0 refers to the undetected class.  
Each cell is a $5cm \times 5cm$ region in the real world.
Given an RGB-D view, we build the occupancy map and semantic map by projecting semantic segmentation images to a 3D point cloud using the available depth maps and the robot poses, discretizing the point cloud into a voxel grid and taking the top-down view of the voxel grid.
The occupancy map depends on the height of the points projected to each cell. 
The semantic map depends on the majority category of the points located at the top grid of each cell.

\begin{figure}[t]
\centering
\includegraphics[width=1.0\textwidth]{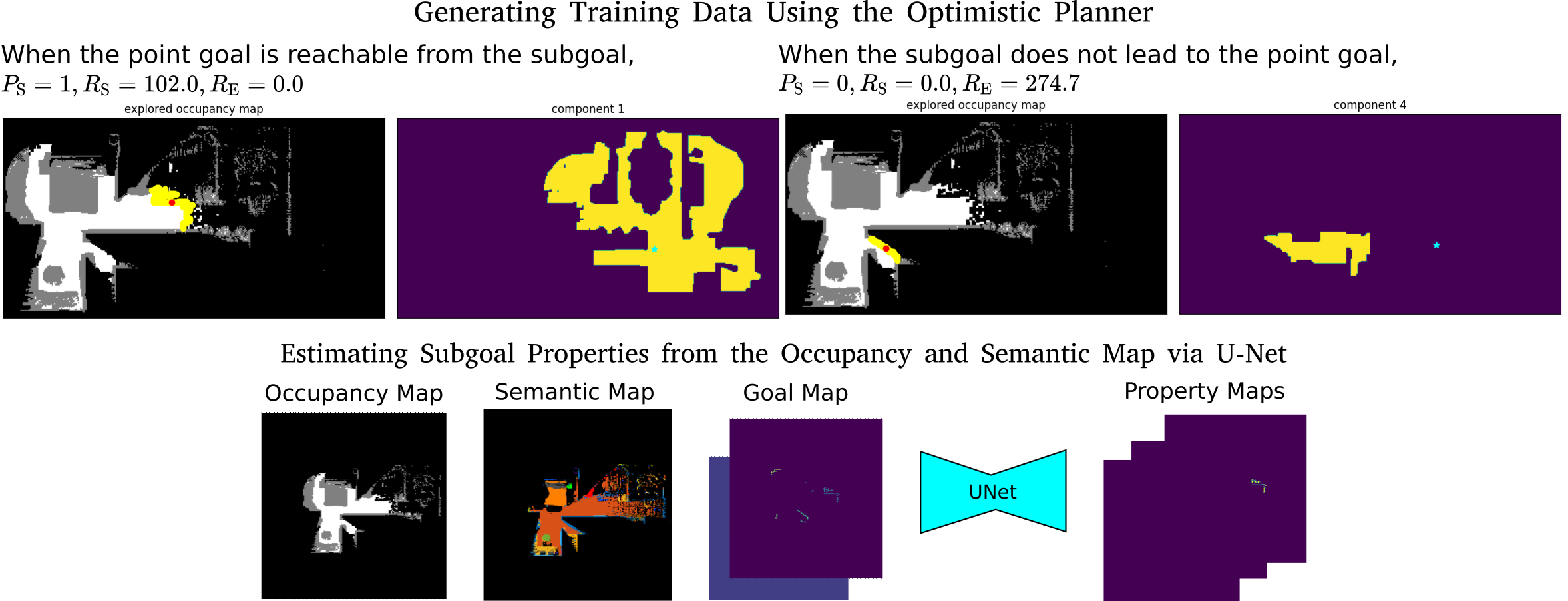}
\caption{We generate training data by having an optimistic planner navigating the environment, maintaining an occupancy and semantic map, extracting the frontiers from the map, and computing their properties.
The U-Net model takes in the local occupancy map, local semantic map, and the goal map and estimates properties for all the visible frontiers.}
\label{fig:unet}
\end{figure}

\noindent\textbf{Frontier Computation}\quad{}
Frontiers are the boundary between unknown and free space on the partial map.
A frontier is computed as a set of connected cells on the boundary that follow an eight-neighbourhood connection.
The center of the frontier is the centroid of the points belonging to that frontier.
We further reduce the number of frontiers by filtering out frontiers unreachable from the robot on the current occupancy map.
Fig.~\ref{fig:unet} shows the computed frontiers on the occupancy map.

\noindent\textbf{Local Navigation}\quad{}
The local navigation module uses the Fast Marching Method~\cite{sethian1996fast} to plan a path to the subgoal or the point goal from the current location on the occupancy map. 
It takes deterministic actions (move forward by 25cm, turn left/right by 30 degrees) to reach the subgoal. 
We update the map after each step and replan. 

\section{Learning-Based Modules}
\label{sec:learning}

\subsection{Training Data Generation}
To plan via Eq.~\ref{eq:dp}, we must estimate the values of the subgoal properties $P_\text{S}$, $R_\text{S}$ and $R_\text{E}$. We do so via a neural network, trained with around 230k frames of data, compared to the 2.5 billion frames required to train the state-of-the-art DD-PPO.
To generate training data, we use an optimistic planner with an RGB-D camera sensor to navigate from random start and goal positions in roughly 5,000 trials in each of the 61 environments.
As the robot moves, revealing new portions of the environment, it maintains occupancy and semantic maps, extracts the newly updated frontiers, and computes labels of $P_\text{S}$, $R_\text{S}$ and $R_\text{E}$ used to train a neural network from the underlying ``true'' map of the environment, known to the robot at training time.

If a frontier leads to the goal through the unknown space, $P_\text{S}$ is labeled 1, and $R_\text{S}$ is calculated as the path length to the goal via the unknown space.
If a frontier does not lead to the goal through the unknown space, $P_\text{S}$ is labeled 0, 
and $R_\text{E}$, the traveling distance to explore beyond a frontier and coming back, is computed on the skeleton of the ground truth occupancy map on the unknown space.
For each subgoal (each corresponding to a frontier), we take the neighboring connected component on the unknown region and generate a skeleton of the surrounding ground truth free space. 
To compute $R_\text{E}$, we compute the subgraph of the skeleton and run a TSP solver (Travelling Salesman Problem solver) on the subgraph to get the trajectory length of visiting every node once and returning to the frontier.

We label the calculated values $P_\text{S}$, $R_\text{S}$ and $R_\text{E}$ on frontier pixels.
In the end, we achieve three maps $m_{P_\text{S}}$, $m_{R_\text{S}}$ and $m_{R_\text{E}}$ only having values on the frontier pixels.
For the convenience of computation, we create three masks, $\text{mask}_{P_\text{S}}$, $\text{mask}_{R_\text{S}}$ and $\text{mask}_{R_\text{E}}$, to denote if the pixel has the ground truth label or not.
To indicate the position of the point goal, we create a 2-channel goal map $m_{q_G}$ by computing the relative coordinates between the point goal and each frontier centroid.
For each frame, we collect the data tuple $(m_t, m_{q_G}, m_{P_\text{S}}, m_{R_\text{S}}, m_{R_\text{E}}, \text{mask}_{P_\text{S}},\text{mask}_{R_\text{S}}, \text{mask}_{R_\text{E}})$ for training the learning module as described in the section below.

\subsection{Learning Module Architecture}
We use a U-Net~\cite{ronneberger2015u} architecture that takes the current observed map $m_t$ and the goal map $m_{q_G}$ as input and outputs a same-sized map with three channels.
Each channel of the output map represents the predictions of $P_\text{S}$, $R_\text{S}$, and $R_\text{E}$ at each pixel.
We train the U-Net through supervised learning.
We add a sigmoid activation after the output of $P_\text{S}$ and use a Binary Cross-Entropy loss $\mathcal{L}_\text{BCE}$ to train it.
The output of $R_\text{S}$ and $R_\text{E}$ is trained with the L1-loss $\mathcal{L}_{\text{L}_1}$. The loss for $R_\text{S}$ is only applied when ground truth $P_\text{S}$ is 1 and the loss for $R_\text{E}$ is only applied when ground truth $P_\text{S}$ is 0. 
The three mask values indicate the positions for which the losses are calculated.
Our final objective is:
\begin{equation}
\label{eq:loss}
\mathcal{L}_{\text{Total}} = \mathcal{L}_\text{BCE} + \lambda * \mathcal{L}_{\text{L}_1}
\end{equation}
These losses are computed only on the frontier pixels indicated by the generated masks.
$\lambda$ is setup as $10^{-2}$ in practice.

Fig.~\ref{fig:unet} shows the data collection process and the U-Net architecture.

\section{Experiments}
\label{sec:exp}

\noindent \textbf{Datasets + Simulated Environments} We use the Habitat~\cite{savva2019habitat} simulator with the Matterport (MP3D)~\cite{chang2017matterport3d} dataset for our experiments. 
MP3D consists of 90 scenes that are 3D reconstructions of real-world environments, and semantic segmentations (40 semantic categories) of egocentric views are available.
We split the MP3D scenes into train:val:test following the standard 61:11:18 ratio to train the perception models and evaluate the navigators.
We compare different Reinforcement Learning approaches using the standard 1008 testing episodes generated for the Habitat PointGoal task~\cite{savva2019habitat}.
For the experiments below, noise-free poses and ground truth egocentric semantic segmentations are provided by the simulator. 

\noindent \textbf{Task Setup} The robot observation space consists of RGB, depth, and semantic segmentation images.
Even though the LSP approach is supposed to perform better with a panoramic camera sensor, to compare with the existing approaches, our agent is equipped with an egocentric $90^{\circ}$ FoV sensor with observations of $256 \times 256$ resolution.
The action space contains four short-term actions: \texttt{MOVE\_FORWARD} by 25cm, \texttt{TURN\_LEFT} and \texttt{TURN\_RIGHT} by $30^{\circ}$ and \texttt{STOP}.

We follow the point goal task setup from~\cite{anderson2018evaluation} where given a goal location, the agent needs to navigate to the goal and stop within a 0.2m radius.
The agent is allowed to move 500 steps to reach the goal.
We use the standard metrics for evaluation~\cite{anderson2018evaluation}; \textit{Success}: percentage of successful episodes,
\textit{SPL}: success rate normalized by path length,
and \textit{SoftSPL}: success rate normalized by the progress to the goal.
 
\noindent \textbf{Implementation Details} 
For the training of the U-Net, we pre-compute 230,000 train / 2,000 val data tuples.
We train the model using PyTorch for six epochs and use the Adam optimizer with a learning rate of 0.001 decayed by a factor of 10 after every two epochs.

\begin{table*}[t]
\caption{Comparison of Point Goal Navigation Methods}.
\label{tab:point_goal}
\centering
\begin{tabular}{l|c|c|c|c|c|c|c|c}
Dataset & \multicolumn{3}{c|}{MP3D Test} \\ 
\hline
Method & Success ($\%$) & SPL ($\%$) & SoftSPL ($\%$) \\ 
ANS~\cite{chaplot2020learning} & 59.3 & 49.6 & - \\ 
DD-PPO~\cite{wijmans2019dd, ramakrishnan2021habitat} & \textbf{89.0} & \textbf{80.0} & - \\
Optimistic~\cite{yamauchi1997frontier} & 75.9  & 60.1 & 64.2 \\
LSP-UNet~\cite{stein2018learning}  & 82.0 & 60.5 & 63.4 \\ 
\hline
Optimistic*~\cite{yamauchi1997frontier} & 87.5  & 70.5  & 72.7 \\
LSP-UNet*~\cite{stein2018learning}  & 88.2  & 74.7  & 77.0 \\ 
\end{tabular}
\end{table*}

\begin{figure}[t!]
\centering
\includegraphics[width=1.0\textwidth]{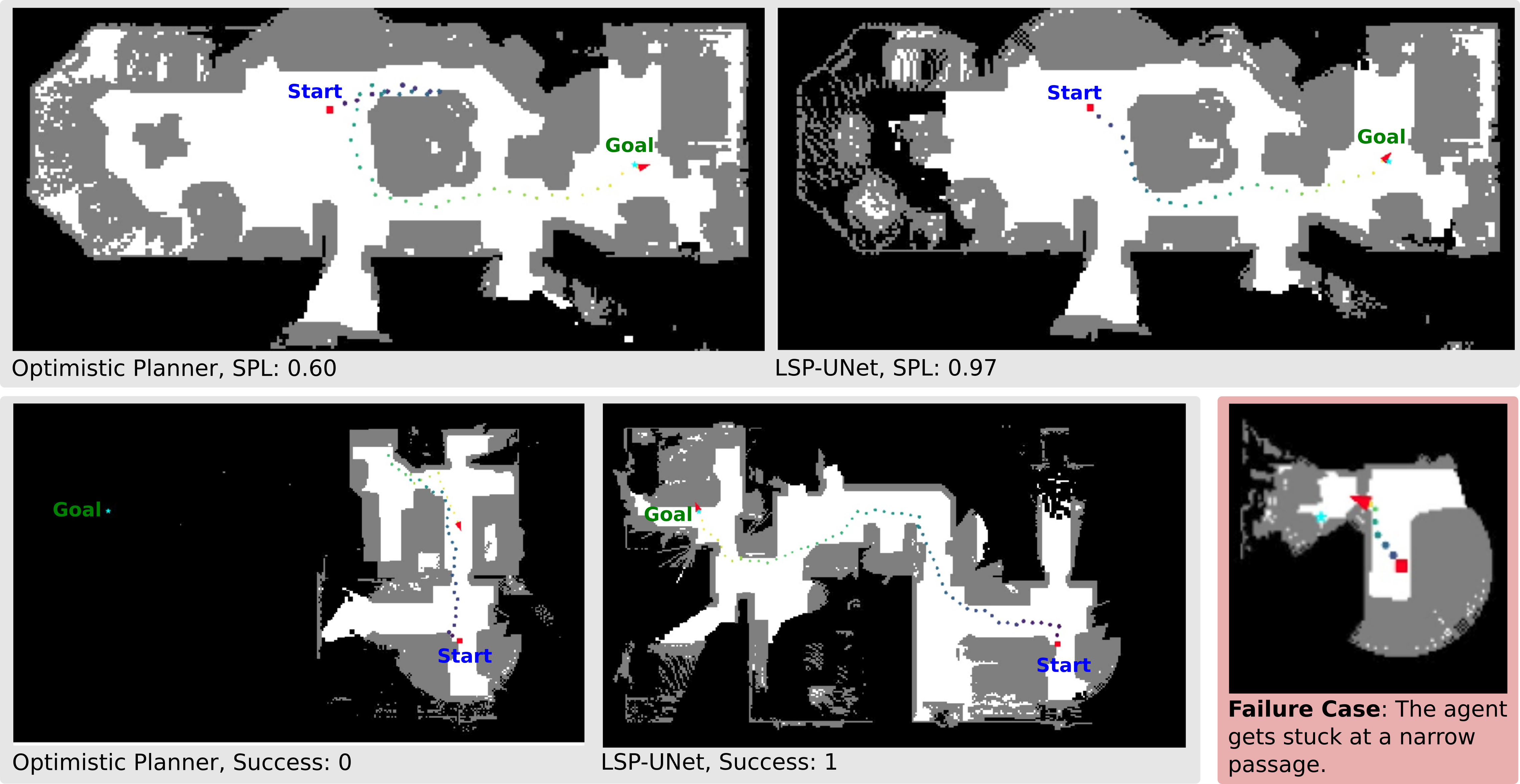}
\caption{Success and failure examples running the \textit{Optimistic} and \textit{LSP-UNet} planner. 
The \textit{LSP-UNet} planner generally knows to avoid dead ends and achieves better-expected performance than the \textit{Optimistic} planner.
The typical failure cases are the fault of the local planner.
}
\label{fig:test_cases}
\end{figure}

\subsection{Point Goal Navigation with an Egocentric $90^{\circ}$ FoV Sensor}
We compare the LSP-UNet approach with the following three methods:\\
\noindent\textbf{Active Neural SLAM (ANS)}~\cite{chaplot2020learning}: A learned exploration model trained with model-free Deep RL using coverage as rewards.
We use this model as an example for modular model-free RL approaches.
It has a global policy module for selecting long-term goals given the currently observed occupancy map and agent-visited location history as input.
We use the results evaluated on MP3D from the original paper~\cite{chaplot2020learning}.\\
\noindent\textbf{DD-PPO}~\cite{wijmans2019dd}: DD-PPO essentially solves the point goal navigation task on the Gibson Environments.
We include it as the example work for end-to-end model-free RL models.
We report the numbers using their latest results trained on the MP3D scenes~\cite{ramakrishnan2021habitat}.\\
\noindent\textbf{Frontier-Based Optimistic Planner (Optimistic)}~\cite{yamauchi1997frontier}: A non-learned greedy algorithm -- which plans by assuming all unknown space is unoccupied and uses the path length from the agent to the point goal through each subgoal as the heuristic.
We use the same optimistic planner for data generation in Sec.~\ref{sec:learning}.

We show the evaluation results in Table~\ref{tab:point_goal}.
The \textit{Optimistic} and \textit{LSP-UNet} methods outperform \textit{ANS} by over $15\%$ in \textit{Success} but are at least $7\%$ worse than \textit{DD-PPO}.
We observe that because the implemented mapping module (described in detail at Sec.~\ref{sec:map}) does not involve learning, it is prone to the perception noise from the simulator.
It sometimes misrecognizes free locations as occupied, which dramatically affects the performance of the entire navigation framework.
To reduce the effect of the mapper, we further evaluate \textit{Optimistic} and \textit{LSP-UNet} with an oracle mapper that produces occupancy maps with no errors.
We name the two approaches as \textit{Optimistic*} and \textit{LSP-UNet*} in Table~\ref{tab:point_goal}.
The performance of \textit{LSP-UNet*} is similar to \textit{DD-PPO} but slightly worse in \textit{SPL}.
\textit{DD-PPO} performs the best in the existing approaches without an explicit map. 
We analyze the failure cases of \textit{Optimistic*} and \textit{LSP-UNet*} and illustrate some example success and failure cases in Fig.~\ref{fig:test_cases}.
A failure case usually happens when the local navigation module has difficulty passing through a narrow passage.
Considering that the typical failure cases are the fault of the local planner, improving the local planner will be the subject of future work.
We can use A* with motion primitives instead.

\section{Conclusion}

We adapted an existing model-based point goal navigation approach, augmented its learning module with a deep convolutional neural network, and compared its performance with a set of SOTA RL-based approaches.
Exploiting advances in visual sensing, semantic mapping, and availability of large datasets of rendered 3D models of real indoor scenes, we showed how to train a deep convolutional neural network to learn three informative properties of the frontiers: (i) likelihood of the frontier leading to the goal, (ii) expected cost of reaching the goal via the frontier and (iii) expected cost of exploring the frontier.
Our experiments suggest that our model is comparable to DD-PPO in point goal navigation in performance and requires much less training data. Another desirable property is that the representations used by our approach, e.g., semantic map, are reusable for other navigation tasks.

\clearpage


\bibliography{example}  

\end{document}